\documentclass[sigconf]{acmart}

\AtBeginDocument{%
  \providecommand\BibTeX{{%
    \normalfont B\kern-0.5em{\scshape i\kern-0.25em b}\kern-0.8em\TeX}}}

\setcopyright{acmcopyright}
\copyrightyear{2023}
\acmYear{2023}
\acmDOI{10.1145/1122445.1122456}

\acmConference[Austin '23]{Austin '23: ACM The Web Conference}{April 30--May 04, 2023}{Austin, TX}
\acmBooktitle{Austin '23: ACM The Web Conference,
  30--May 04, 2023, Austin, TX}
\acmPrice{15.00}
\acmISBN{978-1-4503-XXXX-X/18/06}

\usepackage[export]{adjustbox}
\usepackage{multirow}
\usepackage{balance}

\begin{document}

\title{Spoken Language Understanding for Conversational AI: \\Recent Advances and Future Direction}

\author{Soyeon Caren Han\textsuperscript{1}, Siqu Long\textsuperscript{2}, Henry Weld\textsuperscript{2}, Josiah Poon\textsuperscript{1}\\
The University of Sydney\\
Sydney, NSW, Australia\\
\textsuperscript{1}\{caren.han, josiah.poon\}@sydney.edu.au, \textsuperscript{2}\{slon6753, hwel4188\}@uni.sydney.edu.au}


\renewcommand{\shortauthors}{S Han, et al.}

\begin{abstract}
When a human communicates with a machine using natural language on the web and online, how can it understand the human's intention and semantic context of their talk? This is an important AI task as it enables the machine to construct a sensible answer or perform a useful action for the human. Meaning is represented at the sentence level, identification of which is known as intent detection, and at the word level, a labelling task called slot filling. This dual-level joint task requires innovative thinking about natural language and deep learning network design, and as a result, many approaches and models have been proposed and applied.

This tutorial will discuss how the joint task is set up and introduce Spoken Language Understanding/Natural Language Understanding (SLU/NLU) with Deep Learning techniques. We will cover the datasets, experiments and metrics used in the field. We will describe how the machine uses the latest NLP and Deep Learning techniques to address the joint task, including recurrent and attention-based Transformer networks and pre-trained models (e.g. BERT). We will then look in detail at a network that allows the two levels of the task, intent classification and slot filling, to interact to boost performance explicitly. We will do a code demonstration of a Python notebook for this model and attendees will have an opportunity to watch coding demo tasks on this joint NLU to further their understanding.
\end{abstract}




\maketitle

\section{Significance of this Tutorial}
The efficacy of virtual assistants becomes more important as their popularity rises. Central to their performance is the ability of the electronic assistant to understand what the human user is saying to act or reply in a way that meaningfully satisfies the requester. The human-device interface may be text-based, but it is now most frequently voice and will probably include images or videos in the near future. To put the understanding of human utterances within a framework, within the natural language processing (NLP) stack lies spoken language understanding (SLU). SLU starts with automatic speech recognition (ASR), the task of taking the sound waves or images of expressed language and transcribing them to text. Natural language understanding (NLU) then takes the text and extracts the semantics for use in further processes - information gathering, question answering, dialogue management, request fulfilment, and so on. The concept of a hierarchical semantic frame has developed to represent the levels of meaning within spoken utterances \cite{jeong2008triangular}. At the highest level is a domain, then intent and then slots. The domain is the area of information the utterance is concerned with. The intent (a.k.a. goal in early papers) is the speaker's desired outcome from the utterance. The slots are the types of words or spans of words in the utterance that contain semantic information relevant to fulfilling the intent. An example is given in Table \ref{table:frameeg} for the domain \textit{movies}. Within this domain, the example has intent \textit{find\_movie}, and the individual tokens are labelled with their slot tag using the inside-outside-beginning (IOB) tagging format. The NLU task is thus extracting the semantic frame elements from the utterance. NLU is central to devices that desire a linguistic interface with humans; conversational agents, instruction in vehicles (driverless or otherwise), Internet of Things (IoT), virtual assistants, online helpdesks/chatbots, robot instruction, and so on. Improving the quality of the semantic detection will improve the quality of the experience for the user, and from here NLU draws its importance and popularity as a research topic.

\begin{table}[t]
\small
\caption{An example of an utterance as a semantic frame with domain, intent and IOB slot annotation \cite{hakkani2016multi}}
\label{table:frameeg}
\begin{tabular}{|l|l|l|l|l|l|l|}
\hline
\textbf{query}  & find & recent & comedies & by & james & cameron  \\ \hline
\textbf{slots}  & O    & B-date & B-genre   & O  & B-director  & I-director\\ \hline
\textbf{intent} & \multicolumn{6}{c|}{find\_movie} \\ \hline
\textbf{domain} & \multicolumn{6}{c|}{movies}     \\ \hline
\end{tabular}
\end{table}

In many data sets and real-world applications, the domain is limited; it is concerned only with hotel bookings, or air flight information, for example. The domain level is generally not part of the analysis in these cases. However in wider-ranging applications, for example, the SNIPS data set discussed later, or the manifold personal voice assistants which are expected to field requests from various domains, the inclusion of domain detection in the problem can lead to better results. This leaves us with intent and slot identification. What does the human user want from the communication, and what semantic entities carry the details? The two sub-tasks are known as intent detection and slot filling. The latter may be a misnomer as the task is more correctly slot labelling or slot tagging. Slot filling more precisely gives the slot a value of a type matching the label. For example, a slot labelled ``B-city'' could be filled with the value ``Sydney''. Intent detection is usually approached as a supervised classification task, mapping the entire input sentence to an element of a finite set of classes. Slot filling seeks to attach a class or label to each of the tokens in the utterance, making it within the sequence labelling class of problems.

While early research looked at the tasks separately or put them in a series pipeline, it was quickly noted that the slot labels present and the intent class should and do influence each other in ways that solving the two tasks simultaneously should garner better results for both tasks \cite{jeong2008triangular, wang2010strategies}. A joint model that simultaneously addresses each sub-task must capture the joint distributions of intent and slot labels, with respect to the words in the utterance, their local context, and the global context in the sentence. A joint model has the advantage over pipeline models in that it is less susceptible to error propagation \cite{chen2019wais}, and over separate models that there is only a single model to train and fine-tune.

\section{Topics Covered in this Tutorial}
This tutorial presents a comprehensive overview of recent research and development on using Joint Natural Language Understanding for Conversational AI, focusing on chatbots, dialogue systems, and online chats. We first present our vision of using NLP techniques for Conversational Natural Language Understanding to enable and identify user intent and semantic context. Then we introduce three major modules of our tutorial: (1) The Introduction of Joint Natural Language Understanding (Joint-NLU), (2) Joint-NLU: Feature Engineering, (3) Joint-NLU: Main Modelling, and (4) System Demos and Future Directions. The following tutorial topics and their contents are mainly based on the `A survey of joint intent detection and slot filling models in natural language understanding', published in ACM Computing Survey~\cite{acmsurvey} by the tutorial instructors.

\subsection{Joint Natural Language Understanding}
The joint task marries the objectives of the two sub-tasks. As most papers point out, there is a relationship between the slot labels we should expect to see conditional on the intent and vice versa. A statistical view of this is that a model needs to learn the joint distributions of intent and slot labels. The model should also pay regard to the distributions of slot labels within utterances, and one would expect to inherit approaches to label dependency from the slot-labelling sub-task. Approaches to the joint task range from implicit learning of the distribution through explicit learning of the conditional distribution of slot labels over the intent label, and vice versa, to fully explicit learning of the full joint distribution.

\begin{figure*}[t]
  \centering
  \includegraphics[width=0.74\linewidth]{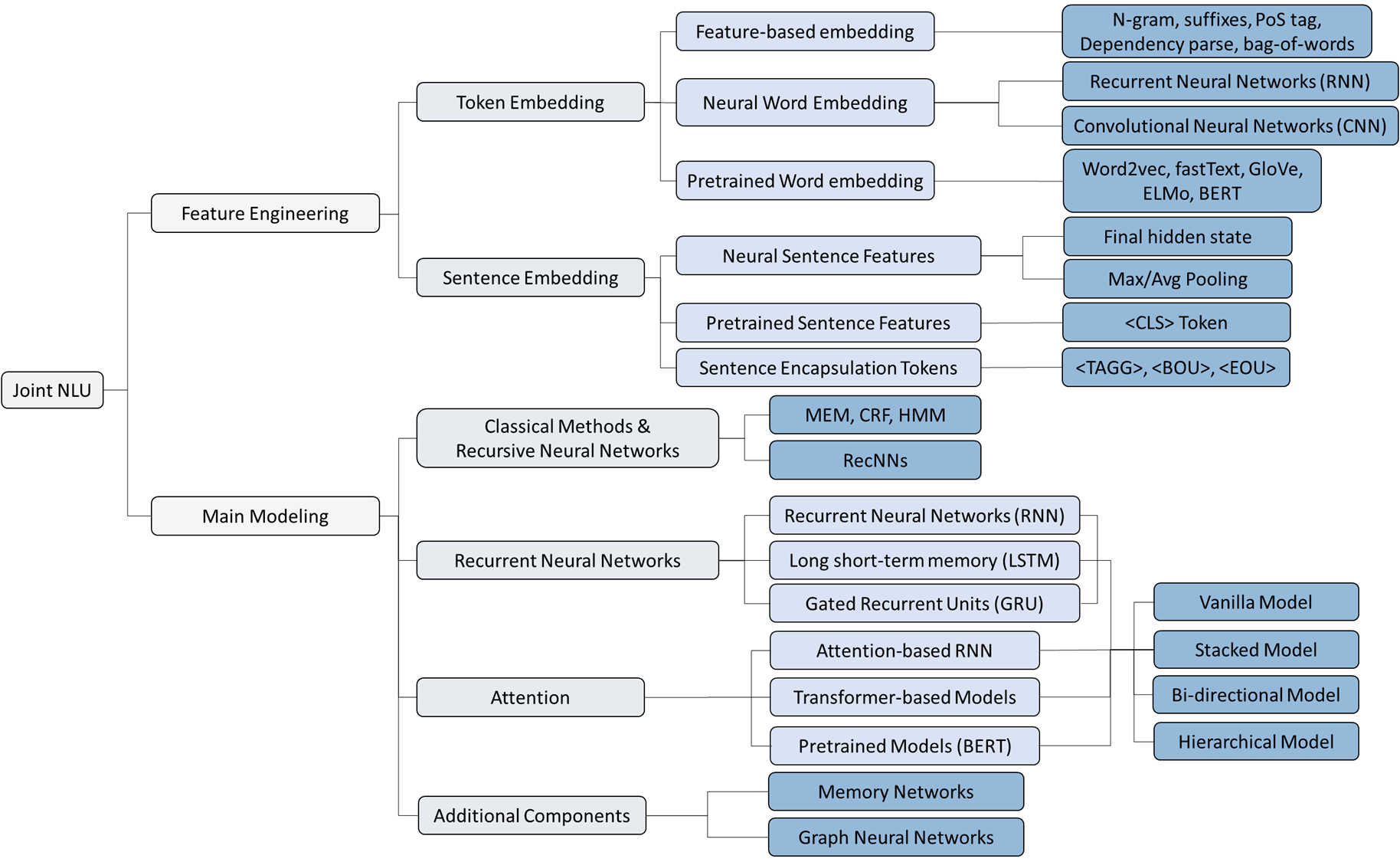}
  \caption{Summary of technological approaches in joint NLU}    \label{fig:overview}
  \Description{Summary of technological approaches in joint NLU}
\end{figure*}

Research in the joint task has largely come from the personal assistant or chatbot fields. Chatbots are usually task-oriented within a single domain, while the personal assistant may be single or multi-domain. Other areas to contribute papers are IoT instruction, robotic instruction (there is also a different concept of intent in robotics to describe what action the robot is attempting), and in-vehicle dialogue for driverless vehicles. These areas also need to filter out utterances not applied to the device. Researchers have also drawn data from question-answering systems, for example, \cite{zhang2016joint} who annotated a Chinese question data set from Baidu Knows. A summary of the technological approaches in joint NLU can be found in Figure \ref{fig:overview}.

\subsection{Joint-NLU: Feature Engineering}
Feature creation is a critical part of the design of circuits in NLU as it ideally should capture, at least, semantic information of the individual tokens, their context, and of the entire sentence. Then, any other extension to the feature set that may be used to enhance the result may be considered, including internal (syntactic, word context) or external (meta-data, sentence context) information.

\paragraph{\textbf{Token embedding}} The earliest models used features familiar from methods like POS tagging and included one-hot word embedding, n-grams, affixes, etc. \cite{jeong2008triangular}. Another approach was to incorporate entity lists from sites such as IMDB (movie titles) or Trip Advisor (hotel names) \cite{celikyilmaz2012joint}. Neural models enable the embedding of diverse natural language without such feature engineering. The gamut of word embedding methods have been used including word2vec (\cite{pan2018multiple, wang2018chinese}), fastText (\cite{firdaus2020deep}), GloVe (\cite{zhang2016joint, liu2019cm, dadas2019deep, okur2019natural, bhasin2019unified, thido2019cross, pentyala2019multi, bhasin2020parallel}), ELMo (\cite{zhang2020graph, krone2020learning}), BERT (\cite{zhang2019joint, qin2019stack, ni2020NLUIoT, han2021bidir, krone2020learning, he2021multitask, huang2021sentiment} and \cite{chen2019bert, castellucci2019multi} (pre-print only)). \cite{firdaus2018deep} and \cite{firdaus2019multi} used concatenated GloVe and word2vec embeddings to capture more word information. A recent approach from intent classification proposed training triples of samples - an anchor sample, a positive sample in the same class and a negative sample from a different class \cite{ren2020siamese}. 

\paragraph{\textbf{Sentence embedding}} The final hidden state in an RNN was used frequently as the sentence embedding  \cite{zhou2016hierarchical, liu2016attention, wang2018chinese}. Sentences were also embedded by using a special token for the whole sentence in \cite{hakkani2016multi, zhang2019using}, as a max pooling of the RNN hidden states \cite{zhang2016joint}, as a learned weighted sum of Bi-RNN hidden states \cite{liu2016attention}, as an average pooling of RNN hidden states \cite{ma2017jointly}, as a convolutional combination of the input word vectors \cite{zhao2018joint, bhasin2019unified}, and as self-attention over BERT word embeddings \cite{zhang2019joint}. Reference \cite{ma2017jointly} also applies a sparse attention mechanism which evaluates word importance over a batch and applies weights within each sample utterance for intent detection.

\subsection{Joint-NLU: Main Modelling}
Multi-task learning has been used by both tasks to look for synergistic learning from other related tasks. The joint task itself is an example of this approach. A joint model needs to learn the joint distributions of intent and slot labels, while also paying regard to the distributions of slot labels within utterances. Approaches to the joint task range from implicit learning of these distributions, through explicit learning of the conditional distribution of slot labels over the intent label, or vice versa, to fully explicit learning of the full joint distributions. The RNN architecture has been exploited as it provides a state at each temporal token step and a final state encapsulating the sentence. One critical observation made of many purely recurrent models was that sharing information between the two sub-tasks is implicit. Attention was used to make this more explicit. Self-attention between the word tokens has been used to learn label dependency and as a stronger alternative to learned weighted sum attention. Transformer encoders are a prevalent non-recurrent architecture that performs self-attention amongst tokens, addresses long-range dependency and can form a sentence representation out of the transformed token representations or by using a special sentence token. This clearly points to the BERT pre-trained model architecture which has been used as input to classifiers and in more integrated architectures. Another method to make influence between the tasks explicit is hierarchical models. Capsule models pass slot deductions of sufficient confidence to an intent detection capsule and vice versa. Memory networks also use a form of explicit feedback. An explicit influence between the two sub-tasks comes even further to the forefront in bi-directional models. Reference \cite{e2019novel} proves slot2intent and intent2slot influence improve results but did not fuse the two approaches. Fusion approaches here include alternating between slot2intent and intent2slot \cite{wang2018bimodel}, post-processing fusion \cite{bhasin2019unified}, and \cite{han2021bidir} in which bi-directional, direct and explicit influence is central to the model architecture. Even within the joint task, some researchers highlight non-optimal handling of label dependency in slot labelling. Adding CRFs to a deep joint model is the common solution to this problem. Graph networks have been designed to garner knowledge from the training data of word-slot, slot-slot, word-intent and slot-intent correlations. The use of sentence context in multi-turn dialogues would seem to provide even more fuel for explicit influence by incorporating intent-to-intent dependency albeit unidirectionally in time through a conversation. Multi-dialogue also offers the interesting, aligned problems of identifying out-of-domain utterances and changes of intent within a conversation.

\subsection{System Demos and Future Directions}
Finally, we conclude our tutorial by demonstrating the system on the publicly available and widely used real dataset (ATIS and SNIPS), how the natural (spoken) language is understood by the two main tasks: Intent detection and slot filling for identifying users' needs from their utterances. All the data, models, codes, and resources will be publicly available. 
We then introduced how it can be applied to online and web chat, including online in-game chat toxicity detection, published by instructors in ACL~\cite{aclconda}. 

\section{Relevance to The Web Conference}
This tutorial is highly relevant to TheWebConf on the topic of web mining and context analysis, focusing on Language technologies and the Web. The tutorial instructors have rich experience in delivering tutorials in major NLP (ACL, COLING, NAACL and InterSpeech), AI (AAAI and IJCAI), web mining and knowledge management (SIGIR, WWW, and CIKM) conferences and journals (ACM Computing Surveys and Knowledge and Information Systems). They also have published in the SLU domain and co-published the joint NLU paper in Interspeech 2021~\cite{interspeechnlu}, NLU-applied papers in ACL 2021~\cite{aclconda} and the joint NLU survey paper in the Journal ACM Computing Surveys~\cite{acmsurvey} since 2021.

\section{Tutorial Outlines}
This tutorial is expected to be 1.5 hour long lecture-style presentation. The lectures will cover the aforementioned topics in great detail while the demo session mainly focuses on providing a clear and practical demonstration of how to set up and implement a basic model for the joint NLU task. The overall outline is listed as follows:

\begin{itemize}
    \item Introduction [20 mins]
        \begin{itemize}
            \item Motivations [5 mins]
            \item Introduction to the NLP and SLU [10 mins]
            \item Unique Challenges of SLU [5 mins]
        \end{itemize}
    \item Joint Natural Language Understanding [10 mins]
        \begin{itemize}
            \item Joint Learning Model in NLU [5 mins]
            \item Explicit and Implicit Joint Learning [5 mins]
        \end{itemize}
    \item Joint-NLU: Architecture [30 mins]
        \begin{itemize}
            \item Joint-NLU: Feature Engineering [15 mins]
            \item Joint-NLU: Main Modelling [15 mins]
        \end{itemize}
    \item System Demo and Future Direction [30 mins]
        \begin{itemize}
            \item Real-world Chatbot Dataset and Models Demo [10 mins]
            \item Online game Chat NLU Demo [10 mins]
            \item Research Problems and Future Directions [10 mins]
        \end{itemize}
\end{itemize}

\section{Previous Editions}
The tutorial is considered a cutting-edge tutorial that introduces the recent advances in an emerging area of using NLU techniques for Conversational AI. Previous ACL, EMNLP, NAACL, EACL, or COLING. TheWebConf tutorials have not covered the presented topic in the past four years. This tutorial has not been presented elsewhere. We estimate that around 60\% of the works covered in this tutorial are from researchers other than the tutorial instructors.

\section{Targeted Audience}
This tutorial is intended for researchers and practitioners in natural language processing, information retrieval, data mining, text mining, graph mining, machine learning, and their applications to other domains. While the audience with a good background in the above areas would benefit most from this tutorial, we believe the presented material would give the general audience and newcomers a complete picture of the current work, introduce important research topics in this field, and inspire them to learn more. Our tutorial is designed as self-contained, so no specific background knowledge is assumed of the audience. However, it would be beneficial for the audience to know about basic deep learning technologies, pretrained word embeddings (e.g. Word2Vec) and language models (e.g. BERT) before attending this tutorial. We will provide the audience with a reading list of background knowledge on our tutorial website.

\section{Tutorial Materials and Equipment}
We will provide attendees with a website with access to all the related information, including the outlines, tutorial materials, references, presenter profiles etc. All the lecture slides will be provided in Google Slides\footnote{\url{https://www.google.com.au/slides/about/}} and the practical demonstration will be conducted via Google Colab~\footnote{\url{https://colab.research.google.com/}}, which is a browser-based and hosted Jupyter notebook service that requires no setup to use. The shareable links to both materials will be available on the website.

\section{Video Teaser}
A video teaser of our tutorial is available on YouTube: \url{https://youtu.be/ovw7093ogeI}.

\section{Organisation Details}
This tutorial will be delivered both in person and online (e.g., via Zoom) during the conference. We will also provide pre-recorded videos as a backup plan that overcomes the potential occurrence of technical problems. We will release our tutorial website and all the materials one week before the tutorial.

\section{Tutorial Instructors}
\paragraph{\textbf{Dr. Soyeon Caren Han}} is a co-leader of AD-NLP (Australia Deep Learning NLP Group) and a Senior lecturer (Associate Professor in U.S. System) at the University of Western Australia and an honorary senior lecturer (honorary Associate Professor in U.S. System) at the University of Sydney and the University of Edinburgh. After her Ph.D.(in 2017), she has worked for 6 years at the University of Sydney. Her research interests include Natural Language Processing with Deep Learning. She is broadly interested in several research topics, including visual-linguistic multi-modal learning, abusive language detection, document layout analysis, and recommender system. More information can be found at \url{https://drcarenhan.github.io/}.

\paragraph{\textbf{Ms. Siqu Long}}\footnote{\url{https://scholar.google.com/citations?user=zeutMxcAAAAJ&hl=en&oi=ao}} is a PhD candidate at the School of Computer Science, University of Sydney. She received her Bachelor's Degree and Master's Degree of Information Technologies in 2016 and 2017 respectively. She worked as a software engineer at IBM, China in 2019. Her research interests include Natural Language Processing and Multi-modal Representation Learning.

\paragraph{\textbf{Dr. Henry Weld}}\footnote{\url{https://scholar.google.com/citations?user=l-u_06gAAAAJ&hl=en&oi=ao}} is a PhD candidate at the School of Computer Science at the University of Sydney. He worked as a data Scientist specialising in machine learning and natural language processing and Senior Quantitative Analyst with over 17 years of front-office experience at the Commonwealth Bank of Australia. His current research interests are in Natural Language Processing. He received B.E. in Civil Engineering from the University of Queensland in 1987 and his first PhD in Pure Mathematics in 1999 from the University of Sydney. He also completed his Master's Degree in Data Science in 2019 at the University of Sydney.

\paragraph{\textbf{Dr. Josiah Poon}} is a co-leader of AD-NLP (Australia Deep Learning NLP Group) and a Senior Lecturer at the School of Computer Science, University of Sydney. He's been using traditional machine learning techniques paying particular attention to learning from imbalanced datasets, short string text classification, and data complexity analysis. He has coordinated a multidisciplinary team consisting of computer scientists, pharmacists, western medicine \& traditional Chinese medicine researchers and practitioners since 2007. He co-leads a joint big-data laboratory for integrative medicine (Acclaim) established between the University of Sydney and the Chinese University of Hong Kong to study medical/health problems using computational tools. \url{https://www.sydney.edu.au/engineering/about/our-people/academic-staff/josiah-poon.html}

\paragraph{\textbf{Specialization in SLU}} These tutorial instructors have worked together in the SLU domain and co-published the joint NLU paper in Interspeech 2021~\cite{interspeechnlu}, NLU-applied papers in ACL 2021~\cite{aclconda} and the joint NLU survey paper in the Journal ACM Computing Surveys~\cite{acmsurvey} since 2021.


\bibliographystyle{ACM-Reference-Format}
\bibliography{sample-base}


\end{document}